\documentclass[runningheads]{llncs}

\usepackage{type1cm}        
\usepackage{makeidx}         
\usepackage{graphicx}        
\usepackage{multicol}        
\usepackage[bottom]{footmisc}
\usepackage{multirow}

\usepackage{newtxtext}       %
\usepackage{newtxmath}       

\makeindex             

\begin{document}

\title{A practical method for occupational skills detection in Vietnamese job listings}

%
%
\author{Viet-Trung Tran\and
Hai-Nam Cao\and
Tuan-Dung Cao}
\authorrunning{V-T Tran et al.}
%
\institute{Hanoi University of Science and Technology, Vietnam
\email{\{trungtv,namch,dungct\}@soict.hust.edu.vn}}


\maketitle

\abstract{Vietnamese labor market has been under an imbalanced development. The number of university graduates is growing, but so is the unemployment rate. This situation is often caused by the lack of accurate and timely labor market information, which leads to skill miss-matches between worker supply and the actual market demands. To build a data monitoring and analytic platform for the labor market, one of the main challenges is to be able to automatically detect occupational skills from labor-related data, such as resumes and job listings. Traditional approaches rely on existing taxonomy and/or large annotated data to build Named Entity Recognition (NER) models. They are expensive and require huge manual efforts. 
In this paper, we propose a practical methodology for skill detection in Vietnamese job listings. Rather than viewing the task as a NER task, we consider the task as a ranking problem. We propose a pipeline in which phrases are first extracted and ranked in semantic similarity with the phrases' contexts. Then we employ a final classification to detect skill phrases. 
We collected three datasets and conducted extensive experiments. The results demonstrated that our methodology achieved better performance than a NER model in scarce datasets.}


\section{Introduction}
\label{sec:1}
Labor market is the foundation and key driver for economic growth. To achieve high efficiency, labor market needs information. Policymakers rely on labor supply and demand relationships to chart economic and social policies. Educators need to align curriculum development with employers' demand, especially in fast-changing sectors. Job seekers need information on skill requirements, company profiles, work conditions, and growth trajectories. In an increasingly digitized economy, labor market policymakers need to continuously maintain an up-to-date vision, focusing on growing skills that are less likely to be replaced by automation~\cite{world2016world}.

It is clear that regulators of Vietnam labor market lack updated and relevant information to make market-driven decisions. Usually, the labor force surveys, which conducted quarterly by General Statistics Office of Vietnam (GSO), face six key challenges: freshness, accuracy, coverage, analysis, usability, and cost. In consequence, Vietnamese labor market has been under imbalance development for many years. Although the number of university graduates is growing, but so is the unemployment rate. In the first quarter of 2018~\cite{le2020competency}, the number of people with intermediate and college degrees that found jobs was 79.1\% and 72.9\% respectively; meanwhile, only 55.6\% of university graduates have jobs.


Besides, Vietnamese job portals have been considered as an important bridge between recruitment managers and job seekers. Over the years, these portals have accumulated a growing amount of digital labor-related market data such as job listings and applicants' resumes. However, the exploitation of these data is limited as these portals only provide job categories and keyword-based search functionality. 


To enable advanced analysis, it is imperative to have a model that can automatically detect skills from labor market-related data. The model can benefit advanced labor market analysis and ultimately facilitate orienting workforce training and re-skilling programs. Various approaches~\cite{zhao2015skill,nadeau2007survey,li2016get,vasudevan2018estimating} consider this skill detection task as a Named Entity Recognition~(NER) task in natural language processing. They have a common drawback: a large number of labeled sentences is needed to train the NER models in a supervised setting. 
Other approaches detect skills from a given document by performing a direct match between n-gram sequences and terms in the target taxonomy~\cite{kivimaki2013graph,bastian2014linkedin,javed2017large}. These approaches, however, do not work for Vietnamese language as there is no such a taxonomy yet. 

In Vietnamese job listing websites, a job opening usually has a common semi-structural format. Each job opening has the following sections: 
\begin{itemize}
\item \textbf{Title}\quad A short, one sentence highlighting for the job to attract job seekers. The title often mentions job position, job level, and salary range.
\item \textbf{Description}\quad One paragraph or a list that describes the job characteristics: What and how the work will be carried on. 
\item \textbf{Compensation}\quad One paragraph or a list that shows salary range and benefits paid to employees in exchange for the services they provide. 
\item \textbf{Requirements}\quad One paragraph or a list that contains experiences, qualifications, and skills necessary for the candidates to be considered for a role. 
\item \textbf{About the company}\quad Brief introduction to the company and its environment. 
\item \textbf{Contact point}\quad An email address and a phone number to submit and question the application. 
\end{itemize}

The order of those sections may vary, however, most skill mentions will be within the requirement section. In this paper, we present a practical approach for skill detection in Vietnamese job listings. Rather than viewing the task as a NER task, we model the task as a ranking problem. Our approach exploits the structural property of a job description: any skill mention found in a requirement section will have a high semantic similarity score with the section itself. 

The rest of this paper is organized as follows: we start in Section 2 by outlining the main steps of the proposed method. In Section 3, we describe in detail the implementation of the tasks in the previous section: embedding, phrase mining, term ranking, and term classification. In Section 4, we carry out a comprehensive experimental study to validate the proposed method. We conclude with a summary of results and future work in Section 5.

\section{Methodology} 
\label{approach}

\begin{figure}[htbp]
\centering
\includegraphics[width=0.40\textwidth]{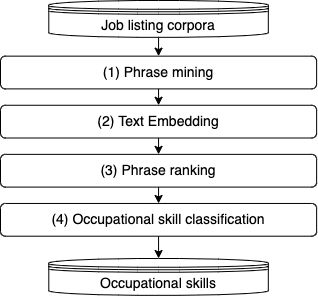}
\caption{The pipeline of our proposed method}
\label{fig:methodology}
\end{figure}

Our method is depicted in Figure~\ref{fig:methodology}. In comparison to the traditional NER approach, our methodology is more practical and less expensive in terms of manual efforts. It is a pipeline composed of 4 layers:
\begin{enumerate}
\item \textbf{Phrase mining}\quad Occupational skill mentions can be multi-word or single-word phrases (e.g., "Java", "data mining"). Thus, a crucial step in our pipeline is phrase mining, which aims at extracting high-quality phrases in a given document. The output of this layer is considered as skill mention candidates for the next ranking layer. To reduce manual effort, we leverage a semi-supervised, weak supervision approach for this layer. 

\item \textbf{Text embedding}\quad This layer is responsible to output the corresponding embedding vectors, given a word, a phrase, a sentence, and a paragraph. The output of the embedding layer will be used to compute ranking similarity scores. For this layer, we can leverage powerful embedding methods such as SIF~\cite{arora2016simple}, and BERT~\cite{devlin2018bert}. Thus, this layer requires no manual labeling effort.   
\item \textbf{Phrase ranking}\quad This layer ranks the importance of a phrase w.r.t. its outer context, the parental requirement section in the job description. Apparently, it can be observed from practical experiments that skill-related phrases often achieve high rankings, while low-ranked phrases are undoubtedly not occupational skills. Therefore, ranking important phrases are capable of discarding clusters of words that are irrelevant to the occupational topic. This layer does not require labeling effort.
\item \textbf{Occupational skill classification}\quad
Generally, there are cases where many extracted phrases in the previous steps are not occupational skill terms. Therefore, this layer is necessary as a binary classification model to identify truly occupational skill terms. This layer requires a labeled dataset. Beginning with a subset of the output phrases in the ranking layer, our labeling workers are required to prepare two subsets: skill and non-skill ones. In contrast, NER labeling workers must carefully select the phrase spans and assign the corresponding labels. Thus, the dataset construction effort is generally cheaper and faster in our methodology than that of the NER approach. 
\end{enumerate}

\section{Implementation} \label{implementation}
\subsection{Phrase mining}

A phrase is defined as a consecutive sequence of words in the text, forming a complete semantic unit in that given context. Phrase mining is the process of high-quality phrases extraction in a given corpus. Phrases can be multi-words and single-word phrases. There are several existing methods for phrase mining, but they suffer from domain dependence, human labeling, and a variety of languages. 

In search of appropriate methods, we select AutoPhrase~\cite{shang2018automated} method for its proven significant effectiveness in different domains. Autophrase only requires a collection of "good" phrases which are cheap to construct from external sources such as Wikipedia or existing dictionaries. Autophrase emphasizes the following features for the detection of high-quality multi-word phrases:


\begin{itemize}
\item \textbf{Popularity}\quad High-quality phrases should be popular. Autophrase uses the probability of occurrence $p(v)$ as the popularity score of $v$ in a given dataset $d$.

For each word or phrase $u$ in a given dataset $d$, the probability of occurrence $p(u)$ is defined as:

\begin{equation*}
{p}(u)=\frac {f_{u}}{\sum _{u'\in d}{f_{u'}}}
\end{equation*}

where $f_u$ is the raw frequency, that is the raw count of $u$ in the dataset $d$.  




\item \textbf{Concordance}\quad For each high-quality phrase, the collocation of its constituent words should occur with a higher probability than what is expected due to chance. Given a phrase $v$, let $u_l, u_r$ denote the two most likely left and right parts of $v$ such that the concatenation of $u_l, u_r$ is equal to $v$ and the Pointwise Mutual Information ($PMI$) of $u_l, u_r$ is minimized:
\begin{equation*} 
u_l, u_r = argmin_{u_l+u_r=v}PMI(u_l, u_r)
\end{equation*}
where the operation $u_l+u_r$ is the concatenation of $u_l$ and $u_r$; $PMI(u_l, u_r)$ is the Pointwise Mutual Information score. $PMI$ is calculated as follows:
\begin{equation*} 
PMI(u_l, u_r) = log\frac{p(v)}{p(u_l)p(u_r)}
\end{equation*}
where $p(v)$, $p(u_l)$, $p(u_r)$ are the probability of occurrence of $v$, $u_l$, $u_r$ respectively.

Another concordance feature can be used is the Pointwise Kullback-Leibler divergence ($PKL$):
\begin{equation*} 
PKL(v||<u_l, u_r>) = p(v)log\frac{p(v)}{p(u_l)p(u_r)}
\end{equation*}

\item \textbf{Informativeness}\quad refers to the possibility of a high-quality phrase being indicative of a specific topic or concept. Informativeness feature includes: (1) Whether stop-words are located within the phrase candidate; (2) Average inverse document frequency (IDF); (3) Probabilities of a phrase in quotes, brackets, or capitalized.

\item \textbf{Completeness}\quad refers to the possibility of a high-quality phrase to express a complete semantic unit in a given document context. 

\end{itemize}

Leveraging these features, Autophrase method of extracting high-quality phrases consists of two main steps:
\begin{enumerate}
    \item Estimate quality of phrases by positive-only distance training.
    \begin{itemize}
        \item A collection of "good" phrases is collected from external sources such as Wikipedia or existing dictionaries. These "good" phrases are put on a positive pool. 
        \item Given the document collection, every $n-gram$ phrase is extracted. Those phrases that satisfy the popularity threshold are treated as candidate phrases. Phrases that are not in the positive pool are put on a negative pool. 
        \item Since training a classifier to distinguish "good" phrases and "bad" phrases directly from the whole noisy data is not effective. Autophrase proposed to use an ensemble method that composes of multiple independent classifiers, each is trained in a subset of mixture data from a positive pool and a negative pool. The output of the trained model is the phrase quality score. 
    \end{itemize}

    \item Re-estimate quality of phrases based on phrasal segmentation to rectify the inaccurate phrase quality initially estimated. 
    \begin{itemize}
    \item This step aims to guarantee the completeness characteristic of detected phrases by exploiting the local context in a given document. Autophrase proposes a POS-guided phrasal segmentation to locate the boundaries of phrases more accurately.
    \end{itemize}

\end{enumerate}

In order to detect salient phrases in Vietnamese corpus, We adapt AutoPhrase to accept the Vietnamese tokenizer and POS tagger. Then, we train Autophrase on the Vietnamese job listing corpus. High-quality phrases are constructed by crawling Wikipedia and job listing websites. 

\subsection{Text embedding: Universal methods for words, phrases, sentences, and paragraphs}

Embedding is one of the most popular representation that has radically transformed the natural language processing (NLP) landscape in recent years. Embedding mechanisms can encode pieces of text, such as word, phrase, sentence and paragraph, as fixed-sized vectors, namely embedding vectors, that allow capturing context, syntactic and semantic similarity. 

Word2vec, Bert are the two most popular word embedding mechanisms. There are dozens of proposed methods to encode sentences, and phrases. One simple method is to average vectors of words in the sentence. Kiros et al.~\cite{kiros2015skip} trains an encoder-decoder architecture to reconstruct the surrounding sentences. InferSent~\cite{conneau2017supervised}, Sentence-Transformer~\cite{reimers2019sentence} learn sentence embeddings by training siamese-based networks with labeled paraphrases. 

In our methodology, we aim at being able to measure the similarity from words, phrases to sentences and paragraphs. So, we use the same embedding mechanisms that can encode various types of text pieces to the same multi-dimensional space. Specifically, we experiment with SIF~\cite{arora2016simple}, SimCSE~\cite{gao2021simcse}, PhoBert~\cite{nguyen2020phobert}, and SBERT~\cite{reimers2019sentence} embeddings. 

\subsubsection{SIF scheme}

SIF (Smooth Inverse Frequency)~\cite{arora2016simple} is a method of averaging
weight to improve sentence embedding performance, with the weight being the inverse frequency smoothed. SIF relies on the assumption that frequent words in the whole dictionary pool are less important than the infrequent words in the computation of the sentence embedding. 

The weights of words in SIF are computed as follows:
\begin{equation*} 
\textrm {Weight(w)} = {\frac {a}{a+f_{w}}}
\end{equation*}

Where $f_w$ is the raw count of the occurrence of the word $w$ in the dataset, $a$ is a hyper-parameter, whose values range from $10^{-3}$ to $10^{-4}$ .
The embedding vector $v_s$ of a sentence $s$ is calculated as follows:
\begin{equation*} 
v_{s} =\frac {1}{\left |{ s }\right |}\sum \limits _{w\in s} {\textrm {Weight(}w)v_{w}}
\end{equation*} 
Where $v_w$ is the word embedding vector of $w$.

SIF works well with most language models pre-trained. Despite the simple formula, SIF works surprisingly well on sentiment tasks, and the tasks of measuring semantic text similarity. The embedding vectors computed by SIF well reflect the intrinsic meaning of words in the sentences. 

\subsubsection{SimCSE scheme}
SimCSE~\cite{gao2021simcse} is a method to train sentence embeddings without having training data. In training, SimCSE encodes a sentence twice by using different dropout masks. A contrastive loss is used to minimize the distance between the two embedding outputs, while minimizing the distance to other embeddings of the other sentences in the same batch (these are negative examples). 

SimCSE serves well as our embedding layer as there is currently no Vietnamese paraphrase datasets. Thus a supervised training is not possible. We leverage pre-trained PhoBert and fine-tune SimCE on a collection of 200K Vietnamese job listings. The trained model is used to predict embedding vectors for all types of text pieces such as words, phrases, sentences, and paragraphs. 

\subsection{Term ranking}

To rank a phrase to its container, the job requirement section, we measure the cosine similarity of their respective embedding vectors $v_p, v_d$. 
\begin{equation*}
SIM(v_p,v_d)=\cos(\theta )=\frac{{v_p} \cdot  {v_d}} { \| {v_p} \|\|{v_d} \|}
\end{equation*}


After obtaining similarity scores of phrases, all phrases with scores lower than a defined threshold (e.g., 0.5) will be discarded since they are definitely irrelevant to skills.

\subsection{Occupational skill classification}

We can select any classification algorithm to implement this layer. In our experiment, we build a small multi-layer neural network for this purpose. The only feature we used for the classification model was term embeddings. 






\section{Experiments and Results} \label{experimentandresults}
\subsection{Data collection}
To demonstrate our methodology, we prepare multiple datasets as follows: 
\begin{itemize}
\item \textbf{Vn\_job\_corpus}\quad Vn\_job\_copus is a collection of 1.122.159 unique sentences extracted from 370.770 Vietnamese-language job listings. The data had been crawled periodically by our crawler system since 2020 Oct. We tokenized the data by vi\_spacy~\footnote{https://github.com/trungtv/vi\_spacy} toolkit. 

\item \textbf{Vn\_requirements\_NER}\quad Vn\_requirements\_NER consists of 625 requirement paragraphs of the job descriptions, with manual labeling of skill terms. The data is split in the ratio of 70:30 for train and test sets.

\item \textbf{Vn\_resumes\_NER}\quad Vn\_resumes\_NER consists of 77284 sentences of the experience sections in the resumes. Due to the scarcity of available sources, these resumes belong only to the information technology sector. The dataset is manually labeled for skill terms. It is also split in the ratio of 70:30 for train and test sets.
\end{itemize}

\subsection{Evaluation Methods}
We compare our proposed methodology to the common NER approach for skill terms detection in Vn\_requirements\_NER and Vn\_resumes\_NER datasets. The models are compared in precision, recall, and F1 scores in complete match and partial match assessments. 

The traditional evaluation requires a complete match of the predicted token and the ground-trust token in order to count a True Positive (TP) case. However, partial match assessment~\cite{nadeau2007semi} is valuable in practice for several reasons: (1) A partial match reveals a significant piece of information: the NER model can predict the right entity, even if whose boundary does not perfectly match with that of the actual entity; (2) The ground-trust datasets are not always in high quality, especially for specific domains in crowd-sourcing data labeling. 

\subsection{Experiments}

Since Vn\_requirements\_NER is more generic where Vn\_resumes\_NER clusters in the information technology sector, we trained two separated NER models on the Spacy framework~\footnote{https://spacy.io/api/architectures}. 

We use the Vn\_job\_corpus to train the Autophrase component, the weighs of words in SIF embedding and SimCSE embedding. We also experiment with two more pretrained embedding models such as PhoBert~\cite{nguyen2020phobert} and SBERT~\cite{reimers2019sentence}. For each embedding method, we have the corresponding ranking component and the corresponding classification component as they rely only on the embedding layers. The training data for the classification component is extracted from the NER dataset. Concretely, positive samples are labeled entities whereas negative samples are randomly selected in the corpus. 


To evaluate the effectiveness of different components in our methodology we measure 4 settings:
\begin{itemize}
\item \textbf{M1: Autophrase}\quad In this setting, we only measure the performance of the Autophrase component. We expect this component to result in a high recall score so that it does not filter out good skill terms. 
\item \textbf{M2: Autophrase $\rightarrow$ Ranking}\quad In this setting, the Autophrase output is used as the input to the Ranking component. We expect the ranking component can boost the precision score while maintaining the high recall score. 
\item \textbf{M3: Autophrase $\rightarrow$ Classification}\quad In this setting, the model consists of Autophrase and Classification components. This setting aims to compare to the last setting in order to clarify the role of our ranking component.
\item \textbf{M4: Autophrase $\rightarrow$ Ranking $\rightarrow$ Classification}\quad This setting is the full pipeline of our methodology. In this setting, the output of Autophrase is passed to the ranking component, then the output of the ranking component will be used as the input for the final classification model.
\end{itemize}

\subsection{Results} \label{results}

\begin{table}
\centering
\caption{Performance evaluation on the Vn\_resumes\_NER}
\begin{tabular}{|p{1.5cm}|c|c|c|c|c|c|c|}
\hline
\multirow{2}{*}{Method}                                                                                     & \multicolumn{3}{c|}{Full match}               & \multicolumn{3}{c|}{Partial match}            & \multirow{2}{*}{Embedding} \\ \cline{2-7}
                                                                                                            & Precision     & Recall        & F1\_score     & Precision     & Recall        & F1\_score     &                            \\ \hline
M1                                                       & 0.28          & \textbf{0.72} & 0.40          & 0.35          & \textbf{0.91} & 0.51          & \multicolumn{1}{l|}{}      \\ 
M2                                                       & 0.32          & 0.66          & 0.43          & 0.40          & 0.83          & 0.54          &                            \\ 
M3                                                       & 0.58          & 0.58          & \textbf{0.58} & 0.67          & 0.67          & \textbf{0.67} & \multirow{3}{*}{SIF}       \\ 
M4 & \textbf{0.59} & 0.53          & 0.56 & \textbf{0.68} & 0.62          & 0.65 &                            \\ \hline
M2                                                                   & 0.30          & 0.64          & 0.41          & 0.38          & 0.81          & 0.52          &                            \\ 
M3                                                                   & 0.58          & 0.58          & \textbf{0.58} & 0.67          & 0.67          & \textbf{0.67} & \multirow{3}{*}{PhoBert}   \\ 
M4 & 0.58          & 0.52          & 0.55          & \textbf{0.68} & 0.60          & 0.64          &                            \\ \hline
M2                                                       & 0.29          & 0.60          & 0.39          & 0.36          & 0.77          & 0.49          &                            \\ 
M3                                                       & 0.58          & 0.58          & \textbf{0.58} & 0.67          & 0.67          & \textbf{0.67} & \multirow{3}{*}{SimCSE}    \\ 
M4 & 0.58          & 0.49          & 0.53          & \textbf{0.68} & 0.57          & 0.62          &                            \\ \hline
M2                                                                   & 0.27          & 0.57          & 0.36          & 0.34          & 0.72          & 0.46          &                            \\ 
M3                                                                   & 0.49          & 0.68          & 0.57          & 0.58          & 0.80          & 0.67          & \multirow{3}{*}{SBERT}     \\ 
M4 & 0.48          & 0.55          & 0.51          & 0.56          & 0.64          & 0.60          &                            \\ \hline
NER                                                                                                         & 0.65          & 0.7           & 0.67          & 0.7           & 0.75          & 0.72          & \multicolumn{1}{l|}{}      \\ \hline
\end{tabular}
\label{tab:result_resume_NER}
\end{table}

\begin{table}
\centering
\caption{Performance evaluation on the Vn\_requirements\_NER}
\begin{tabular}{|p{1.5cm}|c|c|c|c|c|c|c|}
\hline
\multirow{2}{*}{Method}                                                                                  & \multicolumn{3}{c|}{Full match}               & \multicolumn{3}{c|}{Partial match}            & \multirow{2}{*}{Embedding} \\ \cline{2-7}
                                                                                                         & Precision     & Recall        & F1\_score     & Precision     & Recall        & F1\_score     &                            \\ \hline
M1                                                        & 0.19          & \textbf{0.74} & 0.31          & 0.21          & \textbf{0.81} & 0.33          &       \\ 
M2                                                        & 0.28          & 0.73          & 0.41          & 0.30          & 0.79          & 0.44          &                            \\ 
M3                                                        & 0.68          & 0.63          & 0.65          & 0.70          & 0.65          & 0.67          & \multirow{3}{*}{SIF}       \\ 
M4& \textbf{0.72} & 0.62          & \textbf{0.67} & \textbf{0.74} & 0.64          & \textbf{0.68} &                            \\ \hline
M2                                                        & 0.21          & 0.71          & 0.33          & 0.23          & 0.78          & 0.36          &                            \\ 
M3                                                        & 0.68          & 0.63          & 0.65          & 0.70          & 0.65          & 0.67          & \multirow{3}{*}{PhoBert}   \\ 
M4  & 0.68          & 0.61          & 0.65          & 0.70          & 0.63          & 0.66          &                            \\ \hline
M2                                                               & 0.20          & 0.66          & 0.30          & 0.21          & 0.72          & 0.33          &                            \\ 
M3                                                                   & 0.69          & 0.61          & 0.65          & 0.71          & 0.62          & 0.66          & \multirow{3}{*}{SimCSE}    \\ 
M4  & 0.71          & 0.54          & 0.61          & 0.72          & 0.56          & 0.63          &                            \\ \hline
M2                                                                    & 0.20          & 0.68          & 0.31          & 0.22          & 0.74          & 0.34          &                            \\ 
M3                                                                    & 0.67          & 0.60          & 0.63          & 0.69          & 0.62          & 0.65          & \multirow{3}{*}{SBERT}     \\ 
M4  & 0.68          & 0.56          & 0.62          & 0.70          & 0.58          & 0.63          &                            \\ \hline
NER                                                                                                      & 0.32          & 0.24          & 0.27          & 0.37          & 0.27          & 0.31          & \multicolumn{1}{l|}{}      \\ \hline
\end{tabular}
\label{tab:result_requirement_NER}
\end{table}

The experimental results on the Vn\_resumes\_NER and the Vn\_requirements\_NER are depicted in Figure~\ref{tab:result_resume_NER} and Figure~\ref{tab:result_requirement_NER} respectively. Overall, our methodology demonstrates superior performance in comparison to the NER model in the Vn\_requirements\_NER dataset. This result explained that the NER model cannot learn from small labeled data. In the Vn\_resumes\_NER dataset, our methodology achieves lower performance in full match but comparable performance in partial match assessment.

The experimental results shown that despite simple formula, SIF embedding is the best embedding scheme for our embedding component. In Vn\_requirement\_NER dataset, our ranking component based on SIF embedding could filter negative terms so that it boosted up the overall performance of the M4: Autophrase $\rightarrow$ Ranking $\rightarrow$ Classification pipeline. 


\section{Conclusions and Future Work} \label{conclude}
In this paper, we have presented a practical methodology for occupational skills detection in Vietnamese job listings. Our methodology exploits the structural property of a job description: most of the skill mentions are within the requirement section and the skill mentions have high semantic similarity scores with the section itself. According to the best of our knowledge, we are the first to propose such a methodology for skill extraction from Vietnamese text. 

Our methodology is practical so that it does not require expensive manual labeling datasets. Skill mentions are first detected through an automated phrase detection component that relies on limited positive only terms. Then a ranking component based on text embeddings is used to filter out non-related skill terms. The remaining terms are fed to a classification model to finalize the skill detection pipeline.  

Our methodology can achieve comparable performance to a popular industrial-ready NER model in the Vn\_resumes\_NER dataset while being superior in the smaller Vn\_requirements\_NER dataset. 

The future scope of this work is to detect skill synonyms for skill normalization and skill taxonomy construction. This will benefit further analytic works such as job recommendations and labor market analysis.

~\\
\textbf{Acknowledgments.} This research is funded by NAVER corporation.

\bibliographystyle{splncs04}
\bibliography{main}

\end{document}